\documentclass{article}

\usepackage{PRIMEarxiv}
\usepackage{amsmath}
\usepackage[utf8]{inputenc} 
\usepackage[T1]{fontenc}    
\usepackage{hyperref}       
\usepackage{url}            
\usepackage{booktabs}       
\usepackage{amsfonts}       
\usepackage{nicefrac}       
\usepackage{microtype}      
\usepackage{graphicx}       
\usepackage{tabularx}
\usepackage{array}
\usepackage{booktabs}
\usepackage{subcaption}
\usepackage{algorithm}
\usepackage{epstopdf}
\usepackage{algpseudocode} 
\usepackage[table,xcdraw]{xcolor}

\title{A Hybrid Technique for Plant Disease Identification and Localisation in Real-time}

\author{
Mahendra Kumar Gohil\textsuperscript{a,*}, Anirudha Bhattacharjee\textsuperscript{a,1}, Rwik Rana\textsuperscript{b,2}, \\ 
Kishan Lal\textsuperscript{c,3}, Samir Kumar Biswas\textsuperscript{d,4}, Nachiketa Tiwari\textsuperscript{a,5}, Bishakh Bhattacharya\textsuperscript{a,6} \\
\textsuperscript{a}Department of Mechanical Engineering, Indian Institute of Technology Kanpur, Kanpur, Uttar Pradesh, India\\
\textsuperscript{b}Department of Computer Science, University of Washington, Seattle, WA 98105, USA \\
\textsuperscript{c}Department of Plant Pathology, National P.G. College, Gorakhpur, Uttar Pradesh, India \\
\textsuperscript{d}Department of Plant Pathology, C.S.A. University of Agriculture \& Technology, Kanpur, Uttar Pradesh, India \\
\begin{tabular}{@{}l@{}}
\texttt{mgohil@iitk.ac.in, anirub@iitk.ac.in, rwik2000@uw.edu, kishan0885@gmail.com,} \\
\texttt{biswas.csa@gmail.com, ntiwari@iitk.ac.in, bishakh@iitk.ac.in}
\end{tabular}}
\begin{document}
\maketitle
\begin{abstract}
Over the past decade, several image-processing methods and algorithms have been proposed for identifying plant diseases based on visual data. DNN (Deep Neural Networks) have recently become popular for this task. Both traditional image processing and DNN-based methods encounter significant performance issues in real-time detection owing to computational limitations and a broad spectrum of plant disease features. This article proposes a novel technique for identifying and localising plant disease based on the Quad-Tree decomposition of an image and feature learning simultaneously. The proposed algorithm significantly improves accuracy and faster convergence in high-resolution images with relatively low computational load. Hence it is ideal for deploying the algorithm in a standalone processor in a remotely operated image acquisition and disease detection system, ideally mounted on drones and robots working on large agricultural fields. The technique proposed in this article is hybrid as it exploits the advantages of traditional image processing methods and DNN-based models at different scales, resulting in faster inference. The F1 score is approximately 0.80 for four disease classes corresponding to potato and tomato crops.
\end{abstract}


\section{Introduction}
Sustainable agriculture is the key to robust economic growth. Data-driven approaches help to increase agricultural productivity by increasing crop output and reducing losses and the cost of inputs. In most Asian and African countries, such as India, Thailand, and Egypt, the population depends heavily on agriculture for domestic consumption and exports. Hence, improving agricultural productivity is quintessential. One of the significant factors in increasing productivity is the identification of crop diseases and taking preventive steps to restrict their spread to avoid crop loss.

 Image acquisition involves capturing photographic information with the help of an appropriate device. Then pre-processing of captured images is done to improve the image quality. Some image processing methods, such as image filtering, histogram equalization, colour space conversion, and resizing, are examples of the procedures generally deployed at this pre-processing stage to remove noise. Usually, double segmentation is required for high-fidelity tasks such as plant disease recognition. The first segmentation separates the background from the leaf, fruit, or crop. A second segmentation is then conducted to distinguish healthy tissue from the infected ones. Extraction of the feature requires extracting information from the segmented image that could facilitate accurate anomaly detection. The characteristics that may be obtained include texture (energy, contrast, uniformity and similarity), shape, size and colour. Statistical measures such as Colour Co-occurrence Matrix (CCM)\cite{shearer1990plant}, Local Binary Patterns (LBP)\cite{Ahonen2006}, Grey Level Cooccurrence Matrix (GLCM)\cite{Haralick1973}, and Spatial Grey Level Dependence Matrix (SGLDM)\cite{Conners1980} may be used to derive textural features. Model-based methods such as Auto-Regressive (AR)\cite{Ziegel1995} and Markov Random Field (MRF)\cite{Kindermann1980} structures are also used to extract useful features.
 
Sinha et al.\cite{blightdisease}  used statistical texture features derived from Grey-Level-Co-occurrence-Matrix (GLCM) for spot and blight diseases in plant leaves. 

Machine learning algorithms are supplied with feature vectors and trained to classify familiar features associated with each disease. The trained algorithm can then identify characteristics from the new field-captured images. Classification deals with matching one of the classes acquired during training to a given input feature vector. Sanyal et al. \cite{sanyal2013} used colour and textures as features and multilevel perceptron to classify diseases in rice plants. Shao et al.\cite{tobacodisease} proposed multi-feature and genetic algorithms optimizing BP neural network for the automatic identification of tobacco disease. The designer can use multiple learning algorithms to train and identify and merge the algorithm tests \cite{Johannes2017,Es-Saady2016,Padol2017, Debaish}.

As feature extractors and classifiers, DNNs have performed well in image recognition tasks such as the ImageNet challenge in recent years. To accomplish tasks such as disease recognition, insect recognition \cite{Amara2017, Bahrampour2015, Too2019, Picon2019,Mohanty2016,tomatodisease}, weed detection\cite{Bargoti2017},\cite{Ahmad2021}, fruit and flower counting\cite{Bastiaanssen2000,Canziani2017,Chen2017}, as well as fruit sorting and grading\cite{Chen2014}, this principle has been used extensively. Since 2015 agronomists have leveraged deep learning for most research into leaf disease identification using image processing techniques \cite{Christiansen2016}. LeCun et al.\cite{Lecun2015} described deep learning as representation learning, whereby the algorithm finds the best way to represent data through a series of optimizations rather than semantic functions. It is not necessary to do feature engineering with this learning procedure since features are extracted automatically. Andreas and Francesc have presented a thorough survey of deep learning in agriculture \cite{Kamilaris2018}. Deep learning allows advances in disease diagnosis, pest detection, quality management, marketing, automation, robotics and big data in the agricultural sector \cite{Liakos2018}.

Both traditional and deep-learning methods have advantages and disadvantages. While traditional techniques require manual extraction of features and analysis of images, which may contribute to the detection of the false disease. DNN-based approaches have an advantage over traditional methods because they automatically classify useful features. However, currently, DNN is restricted by computational power since images of very high resolution cannot be used for training and validation, as direct usage of the image would increase the training and detection time exponentially. For training, then, images have to be resized which results in the loss of useful features \cite{luke2019} \cite{kannojia2018}. This calls for a hybrid approach in which the advantages of both techniques may be combined.

From the symptoms on the leaves, one can diagnose plant diseases. Different features of the leaves, such as colour, texture and general shape and size may indicate whether the plants are healthy or not. The main objective in identifying the disease is to identify the area in which the characteristics vary from regular healthy leaves. Traditional edge detection techniques like Otsu Segmentation, Watershed Segmentation, Canny edge detection and Sobel Edge Detection may be used. Traditional edge detection techniques may help detect the outline of the individual leaves leading to segmenting an image into individual leaves, followed by disease detection on each leaf. 
 
Many researchers have tried to develop a plant disease detection system and mobile application for plant disease detection. Table \ref{tab:mobile_application} shows recently developed mobile applications.
\begin{table}[]
\caption{Recently developed mobile disease detection applications}
\label{tab:mobile_application}
\begin{tabularx}{\textwidth}{@{}l*{5}{>{\raggedright\arraybackslash}X}@{}}
\toprule
\textbf{Article} & \textbf{Year} & \textbf{Application Name} & \textbf{Plant Disease} & \textbf{Algorithm} & \textbf{Dataset} \\
\midrule
\cite{mobile_majid2013pedia} & 2013 & I-PEDIA & Identification of 4 Specific Diseases in Paddy Crops & Fuzzy entropy and probabilistic neural network (PNN) & Custom Dataset \\ \addlinespace
\cite{mobile_2019idahon} & 2019 & iDahon & Detection of Diseases in Terrestrial Plants & Custom CNN Model & Internet Public Data \\ \addlinespace
\cite{mobile_2022agroaid} & 2019 & AgroAId & Classification of 39 Different Plant Diseases & CNNs (MobileNet, MobileNetV2, NasNetMobile, EfficientNetB0) & Plant Village \\ \addlinespace
\cite{cropcare} & 2020 & CROPCARE & Identification of 26 Specific Diseases across 14 Different Crop Species & MobileNet V2 & Plant Village \\ \addlinespace
\cite{toled} & 2021 & ToLeD & Classification of 9 Tomato Disease Classes & Custom CNN Model & Plant Village \\ \addlinespace
\cite{mobile_2021plantifyai} & 2021 & PlantifyAI & Identification of 26 Specific Diseases across 14 Different Crop Species & Custom Filter + MobileNetv2 & Plant Village \\ \addlinespace
\cite{mobile_2022plantbuddy} & 2022 & PlantBuddy & Identification of 26 Specific Diseases across 14 Different Crop Species & MobileNet V2 and Inception V3 & Plant Village \\ \addlinespace
\cite{MSDNet} & 2022 & MS-DNet & Identification of 16 Disease of Paddy and Corn & Custom CNN Model (MS-DNet) & Plant Village \\
\bottomrule
\end{tabularx}
\end{table}

The remainder of the paper is organized as follows. The proposed methodology is discussed in Section~\ref{section:methadology}.
Section~\ref{section:algorithm} outlines the proposed algorithm. 
Section~\ref{section:result} shows the execution of the suggested approach and implementation. The article is concluded in section~\ref{section:conclusion}.
\subsection{Motivation}
Experts' bare-eye evaluation is the standard plant disease detection and identification approach. However, this needs constant supervision by consultants, which could be outrageously expensive in large farms. Furthermore, farmers may have to travel long distances to reach experts in low-income or developing countries like India, which makes consulting experts expensive and time-consuming. Automated analysis of plant diseases is an essential research topic as it may be beneficial in continuously monitoring large crop fields. This will help in quickly recognizing diseases from the symptoms that appear on the leaves of plants.

\subsection{Contribution}

In this paper, a novel technique for disease identification is proposed. Different features of a particular disease can be detected in a layer-based manner. A quadtree-based layering approach is applied to narrow down the affected area. Each layer detects features in an ordered manner; this would help in getting the region of interest of the diseases on the leaves. Once the area is narrowed down to the requisite level with a preset confidence score, a DNN-based approach is applied for the classification.

\section{Methodology}
\label{section:methadology}
\subsection{Recursive Partitioning of Image }
This step aims to localize the main feature area to detect the disease-prone areas. In general, during neural network training, the images are resized to a smaller size to reduce training time else; if the image size is big, training time may increase exponentially. However, this kind of training has a drawback, i.e. it reduces the feature space. Some features of big images are lost during resizing. A robust plant disease detection algorithm must detect the possible minor feature with good accuracy in minimal time. 

\begin{figure*}[h]
	\centering
	\includegraphics[width=0.8\linewidth]{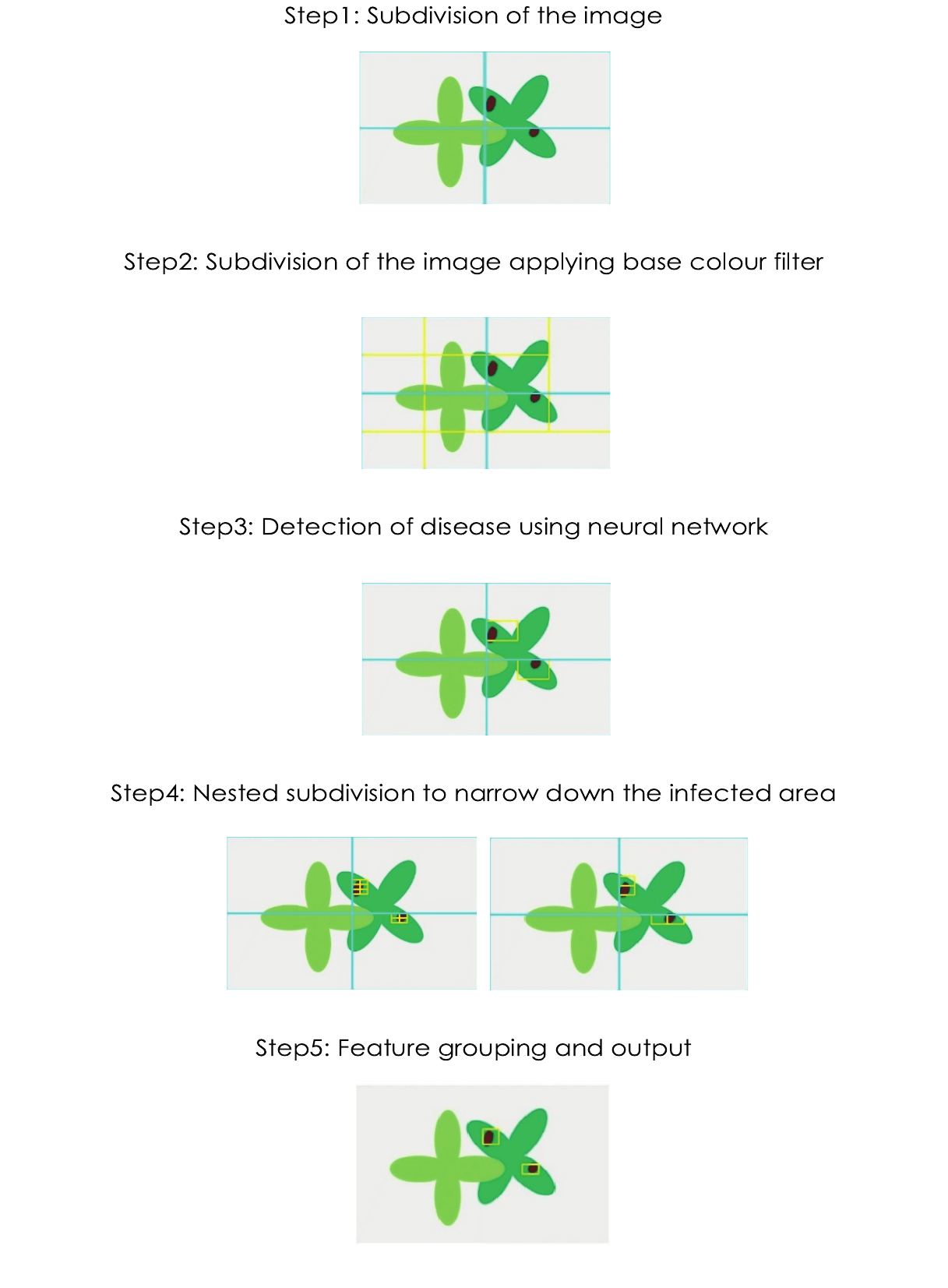}
	\caption{Recursive Partitioning of image.}
	\label{FIG:Recursive}
\end{figure*}

To address this problem, we recursively subdivide the original image to remove undesired feature areas.  Once the feature area is narrowed down, the neural network is applied for disease detection. At this stage, colour is used as a feature to narrow down the disease-prone area. Also, texture information may be used to localize the disease-prone area.
 The proposed partitioning scheme is analogous to quadtree decomposition \cite{Finkel1974} modified with the conditioning schemes, as shown in Figure \ref{FIG:Recursive}.
Quadtree\cite{samet1984quadtree} \cite{Zhou2017}is a very efficient search and data-storing method with an order O(n) search time complexity. A Quadtree of depth 'd' storing a set of 'n' points has O((d + 1)n) nodes and can be constructed in O((d + 1)n) time. This way, the zone of interest can be reached easily in the shortest time possible.

\subsection{Disease Detection Neural Network}
\label{sec:DDNN}
\begin{figure}[ht]
	\centering
		\includegraphics[scale=.2]{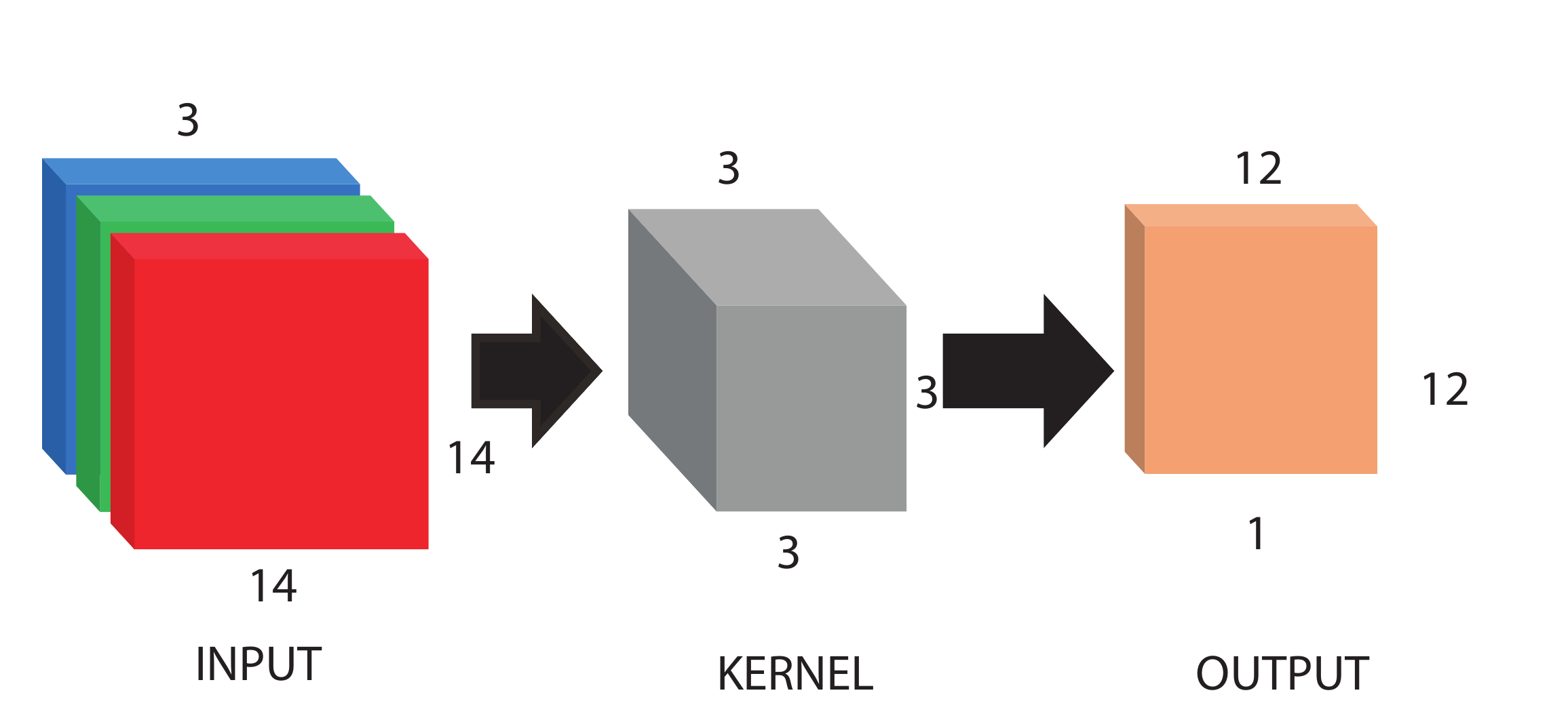}
	\caption{An illustration to demonstrate the functioning of the conventional convolution process applied to a picture input with size 14 × 14 × 3, utilising a single kernel with dimensions 3x3x3}
	\label{FIG:Convolutional}
\end{figure}

\begin{figure}[!ht]
	\centering
		\includegraphics[scale=.35]{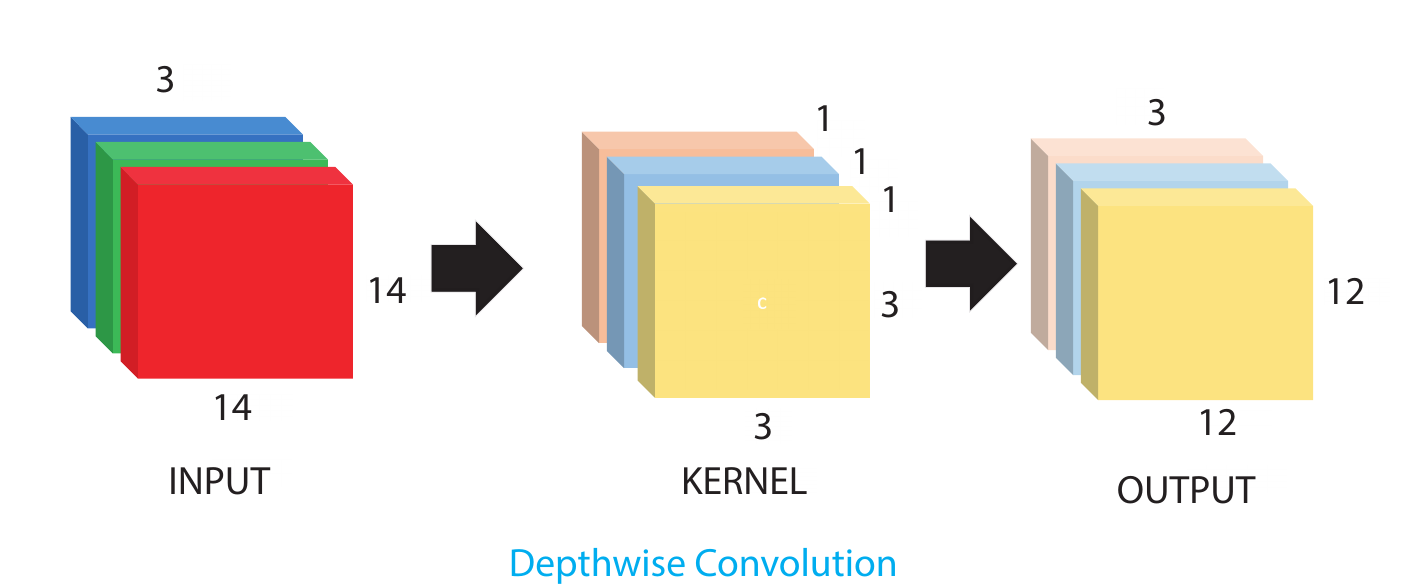}
		\includegraphics[scale=.35]{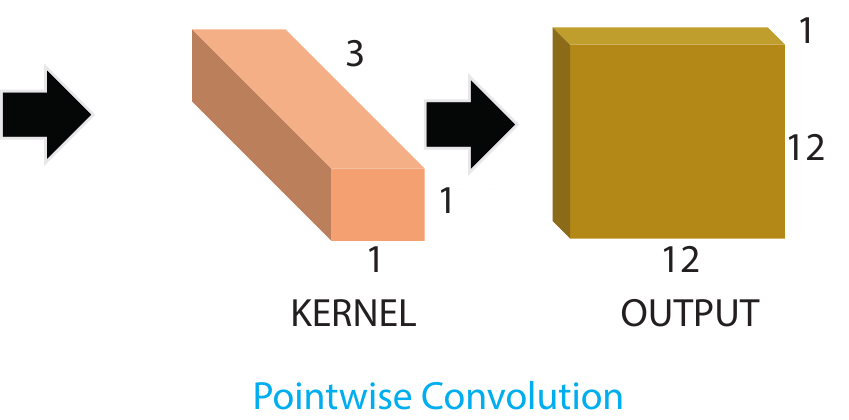}
	\caption{Example of Depth-Wise Separable Convolutions (DWSC) operation using an image with dimensions of 14 by 14 by 3 as input. Three kernels of dimension 3x3x1 are used to perform a depth-wise convolution first, and then one kernel of dimension 1x1x3 is used to perform a pointwise convolution.
}
\label{FIG:DepthandPoint}
\end{figure}

The approach under consideration utilises transfer learning on the Xception neural network architecture to facilitate feature learning in the context of disease detection. François Chollet developed the Xception model in the year 2016 \cite{Chollet2017}. The Xception architecture demonstrated superior performance compared to the VGG-16, ResNet50, ResNet101, ResNet152, and Inception-V3 architectures when evaluated on the ImageNet dataset. 
The foundation of the Xception model is present in previous models like the original Inception \cite{Zeng2016} and the Inception-V3 \cite{Szegedy2016}. The primary distinction of the Xception network lies in the utilisation of Depth-Wise Separable Convolutions (DWSC) instead of conventional Convolutions, along with the incorporation of Residual Connections (RC), which resembles the ResNet models. The architectural design consists of 36 convolutional layers, collectively serving as the foundation for feature extraction. The architecture consists of 36 convolutional layers organised into 14 modules. It is worth noting that the first and last modules are the only ones that lack residual connections. Depth-Wise convolution is a type of spatial convolution that operates on each channel of an input separately. It is then followed by a pointwise convolution, which is a 1 × 1 convolution. This pointwise convolution projects the output of the Depth-Wise convolution onto a new channel space. The visual representation in Figure \ref{FIG:Convolutional} illustrates a conventional convolution process. On the contrary, an exemplification of DWSC is demonstrated in 
Figure \ref{FIG:DepthandPoint}.
The final output of a traditional convolution and DWSC are the same; however, DWSC is more efficient and has lesser time complexity than the traditional methods. Traditional methods involve direct convolution of the kernel over the input data (Figure \ref{FIG:Convolutional}). DWSC, on the other hand, breaks the process into steps.  A single-layer kernel convolves each layer of the input data; hence the output has the same depth as the input. Pointwise convolution is the next stage, which transforms the result of depthwise convolution into the desired result. Deeper models are produced using DWSC and are more effective than wider ones. On the other hand, the Xception model is more effective and quicker than the VGG-16 model despite having many parameters (more than 20 million). 

\begin{gather}
Traditional\_steps = N.Di^2.Dk^2.M \label{eq:1} \\
DWSC\_steps = M.Di^2.(Dk^2 + N) \label{eq:2}
\end{gather}

Equation ~\ref{eq:1} and ~\ref{eq:2} denote the steps of the traditional and DWSC algorithms. $Di$ and $M$ denote the dimension and depth of the input data, respectively. $Dk$ and $N$ denote the dimension and depth of the kernel. From the two equations, it may be noted that DWSC takes lesser steps than traditional convolution. DWSC is more efficient than the traditional algorithm.

\begin{figure*}[h]
	\centering
	    \includegraphics[scale=.37]{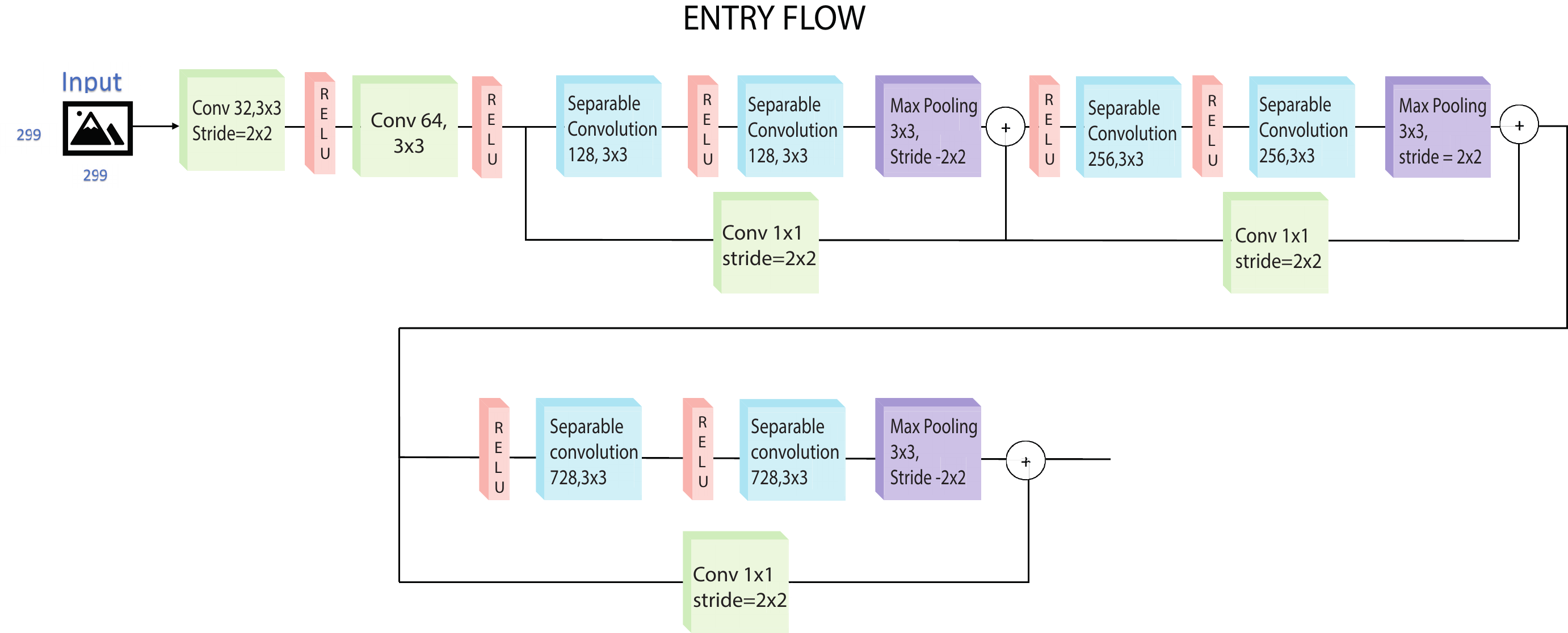}
	    \includegraphics[scale=.08]{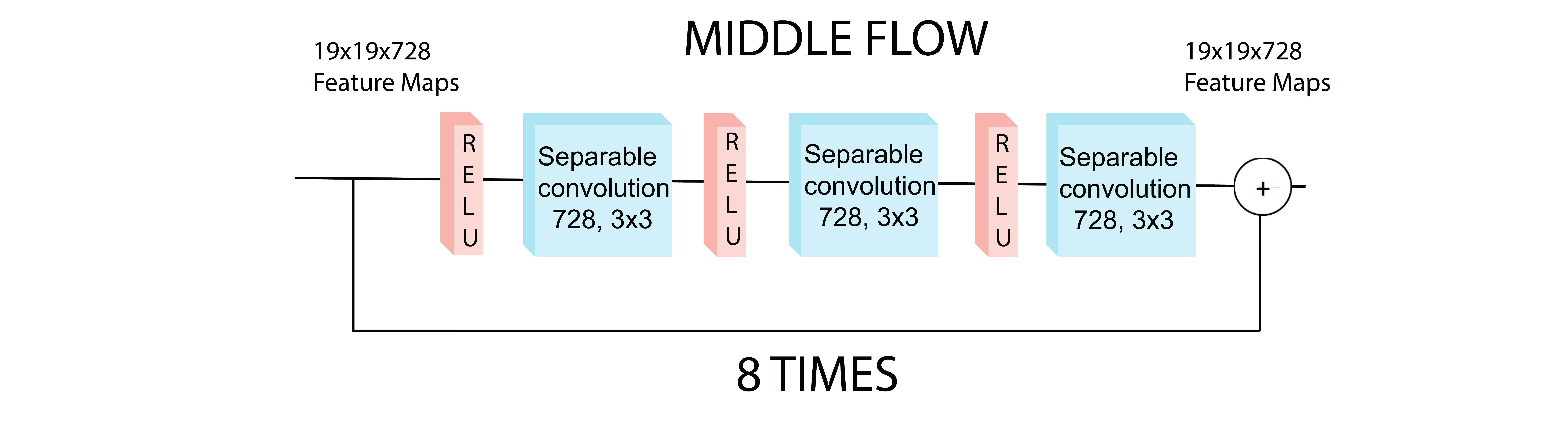}
		\includegraphics[scale=.08]{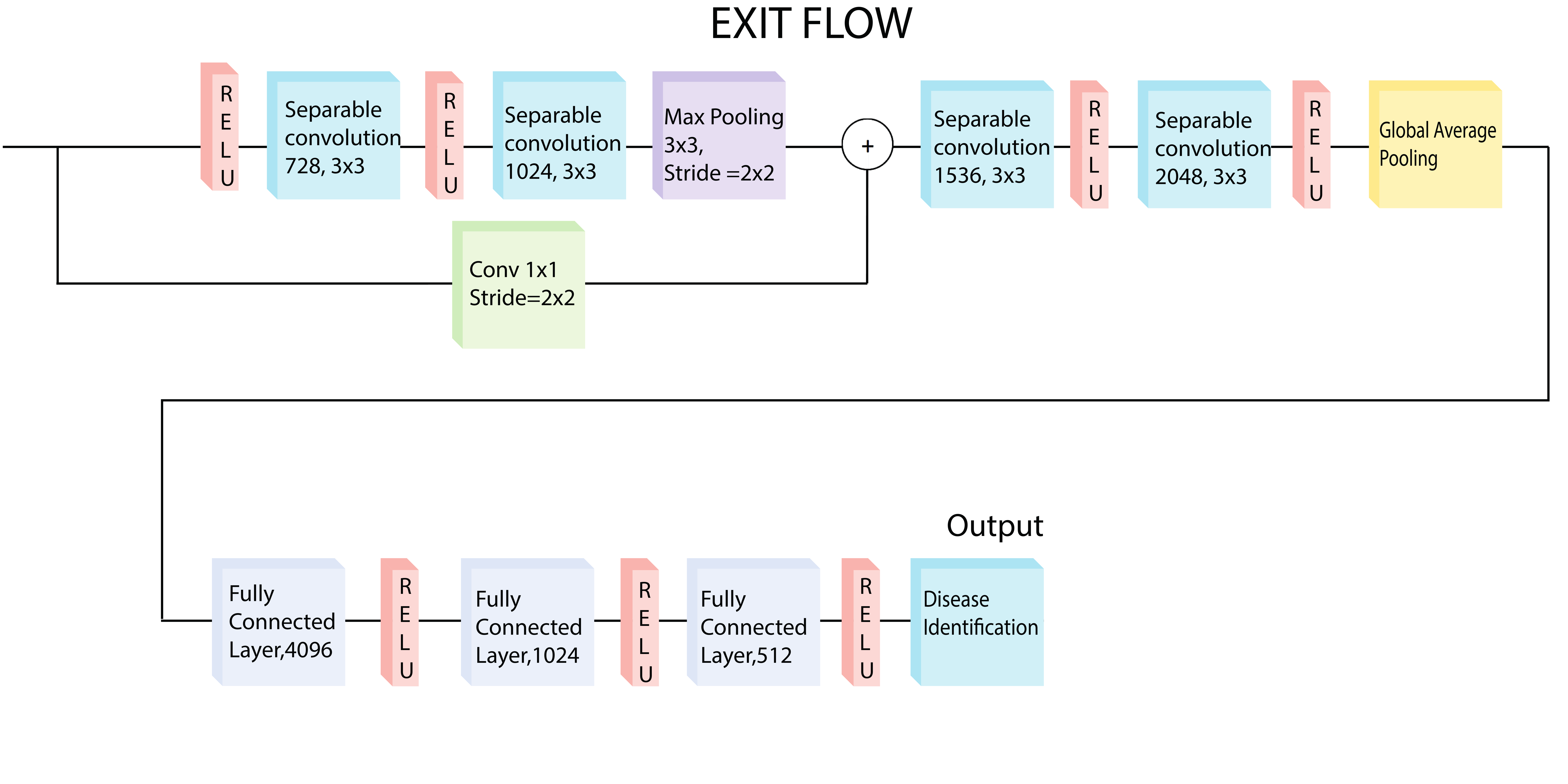}
	\caption{Modified Xception model}
	\label{FIG:xception}
\end{figure*}

Xception modules are diﬀerent from the ones found in the Inception models because of the integration of DWSC.
Figure \ref{FIG:xception} illustrates the modified Xception model used in this work. The last three layers are modified to apply transfer learning for three class predictions. The learning rate in the initial layers is kept low because those layers are already trained in learning primary features. This results in avoiding the requirement of high disturbance. The middle layers have a slightly higher learning rate because they may assist in learning some complex features.  In the new layers, the learning rate is kept at the highest. It helps in learning specific features of the disease.

\subsection{Proposed Hybrid Technique}
The proposed hybrid technique consists of three steps. In the first step, recursive partitioning of the image is carried out to isolate leaves from the background to narrow down the search space. In the next step, a disease-detection neural network is used in the segment for disease identification. Once segments are identified, localization is carried out with the proposed algorithm to delineate affected zones accurately.
\section{Algorithm}
\label{section:algorithm}
\begin{figure}[!ht]
    \centering
    \includegraphics[width=.7\linewidth]{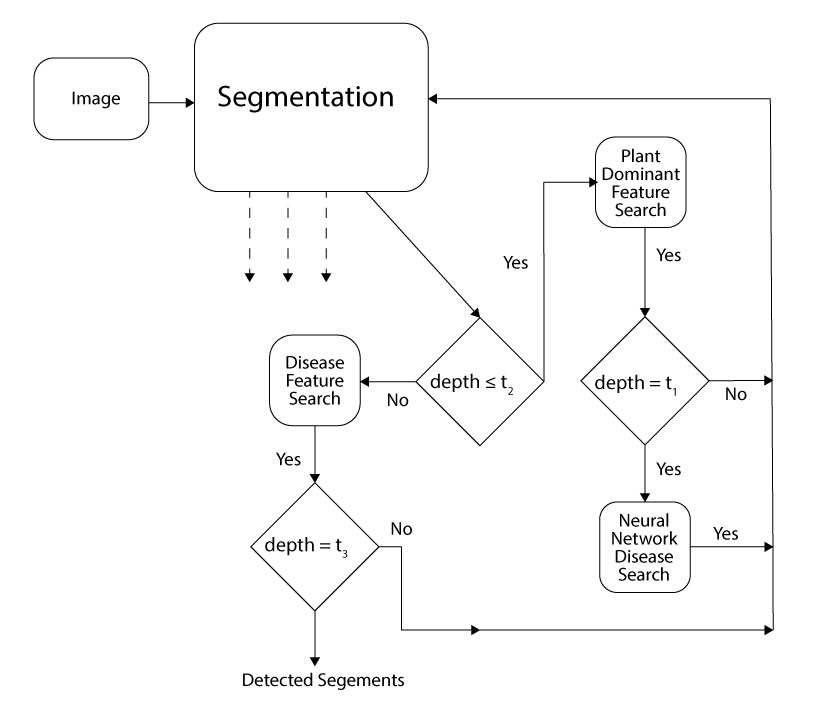}
    \caption{Flow chart of the algorithm}
    \label{fig:flowchart_method}
\end{figure}

The method is inspired by quad-Tree segmentation. 
The algorithm consists of three parts:
\begin{itemize}
\item Segmentation of the image and searching for the disease.
\item Grouping neighbouring segments of the image to get a collective region.
\item Localisation of the affected regions.
\end{itemize}
The flow chart of the algorithm is shown in Figure \ref{fig:flowchart_method}.
\subsection{Segmentation and Disease Feature Detection}
The input image is recursively divided into child segments (each having the same size) as shown in the Recursive Segmentation (Algorithm ~\ref{alg:recursiveseg}). The original image is recursively divided into four parts, and each subsequent part is subdivided into four parts; this process continues, which is analogous to Quad-Tree decomposition. Based upon the feature characteristics (size, colour, texture), the recursive segmentation process continues until a region of interest is identified  or if the QuadTree segmentation reaches the lowest composition of the image, i.e. the size of a pixel. 
\newcommand{\multiline}[1]{%
  \begin{tabularx}{\dimexpr\linewidth-\ALG}[t]{@{}X@{}}
    #1
  \end{tabularx}
}
\begin{algorithm}
    \caption{Recursive Segmentation} 
    \label{alg:recursiveseg}
    \begin{algorithmic}[1]
        \Require DepthLimit, Image, Image Vertices, Depth Count, FeatureDict, Depth Check, SegmentWidth, SegmentHeight
        \If {Depth Check $= 0$ and Depth Count $\neq$ DepthLimit}
            \State FeatureDict $\leftarrow$ \Call{Disease Detection Main}{} \Comment{Update FeatureDict}
            \State Depth Count $\mathrel{+}=$ 1
            \If{SegmentWidth $= 1$ or SegmentHeight $= 1$}
                \State Depth Check $= 1$
            \EndIf
            \State \Call{Recursive Segmentation}{} \Comment{Recursive call to the function}
        \EndIf
        \Ensure FeatureDict, SegmentWidth, SegmentHeight
    \end{algorithmic}
\end{algorithm}

At each layer of the Quad-Tree, specific conditions are imposed with respect to features.   All the segments that do not have a particular feature in a given layer are removed and not segmented further. Initially, Disease Detection Main (Algorithm ~\ref{alg:diseasedetectionmain}) calls Base Colour Detection (Algorithm ~\ref{alg:basecolordetection}), which tries to close down on the leaf region by detecting and locating the green areas in the image.  The segments containing colours other than the specified green range are not selected.
\begin{algorithm}
    \caption{Disease Detection Main}
    \label{alg:diseasedetectionmain}
    \begin{algorithmic}[1]
        \Require DepthLimit, Image, HSV Image, Depth Count, Image Vertices, FeatureDict, SegmentWidth, SegmentHeight, Green Check
        \State Base\_Colour\_Threshold $\leftarrow B$ \Comment{Set $B$ based on input image. $B=0$ for the plant village dataset gives the best results.}
        \State Initialize Features $\leftarrow F_1,F_2,F_3...F_n$ 
        \State Construct a dictionary of the features and base colour limits 
        \Statex Limit\_Dictionary[$F_1$] $\leftarrow B+x_1$
        \Statex \vdots
        \Statex Limit\_Dictionary[$F_n$] $\leftarrow B+x_n$
        \Statex for all $k \in [1,2,...n]$ such that $B+x_k < DepthLimit$
        \State max\_base\_limit $\leftarrow B+max(x_1,x_2,...,x_n)$
        \State FeatureDict, width, height $\leftarrow$ \Call{Base Colour Detection}{} \Comment{Detect base green colour to narrow down plant leaves search in the image.}
        \State FeatureDict, width, height $\leftarrow$ \Call{Disease Feature Detection}{} \Comment{Detect disease-based features after diseases are detected in segments and search is narrowed down using base leaf colour search.}
        \Ensure FeatureDict, width, height
    \end{algorithmic}
\end{algorithm}

After reaching the specified layer, i.e., Base Colour Threshold, in Disease Detection Main (Algorithm \ref{alg:diseasedetectionmain}), Base Colour Detection (Algorithm \ref{alg:basecolordetection}) detects diseases in each of the remaining green segments using the Neural Network Model as described in Section \ref{sec:DDNN}, which is trained on the Plant Village\cite{Hughes2015} and custom dataset collected during field trials from the Chandrashekhar Azad University of Agriculture and Technology, Kanpur. The next step is adding the segments where specific diseases were detected into their corresponding lists in the Feature dictionary (FeatureDict). The disease segments are further subdivided based on the green colour up to a specific layer of the decomposition tree. The optimal depth of this layer in the decomposition tree varies for different disease phenotypes. For the Plant Village dataset, the optimum depth for Early Blight is B+3, and for Late Blight it is B+1, where B (i.e., \texttt{Base\_Colour\_Threshold}) indicates the layer at which disease detection is performed using the neural network architecture of the decomposition tree.
 \begin{algorithm}
    \caption{Base Colour Detection}
    \label{alg:basecolordetection}
    \begin{algorithmic}[1]
        \Require{FeatureDict, Depth Count, Limit\_Dictionary}
        \If{Depth Count $\leq$ Base\_Colour\_Threshold}
            \If{Depth Count $<$ Base\_Colour\_Threshold}
                \State Iterate over all segments in FeatureDict[base colour], check for green colour and add child segments in FeatureDict[base colour]               
            \Else
                \For{segment in FeatureDict[base colour]}
                    \State feature $\leftarrow$ NeuralNetworkModel(segment)
                    \State FeatureDict[feature].append(segment)
                \EndFor
            \EndIf
            \Comment{Note that for 1st layer, the segment would be the entire image}
        \Else
            \For{feature in FeatureDict}
                \If{feature $\neq$ base leaf colour and Depth Count $\leq$ Limit\_Dictionary[feature] }
                    \For{segment in FeatureDict[feature]}
                        \State Break segment into 4 child segments, check for base green colour in each, and append segments with colour above threshold in new coordinates
                    \EndFor
                \EndIf
            \EndFor
        \EndIf
        \State Calculate the width and height of one segment
        \Ensure{FeatureDict, width, height}
    \end{algorithmic}
\end{algorithm}
Once these layers are reached, the segments are filtered concerning disease-specific features using Disease Feature Detection (Algorithm ~\ref{alg:diseasefeaturedetection}). In this work, colour is used as a parameter for disease detection. More feature parameters and their combination can be included at this stage to detect diseases that exhibit a complex composition. This way, a hierarchical search is carried out on the entire image, and the disease segments are filtered.

The segments containing the specific colour of the diseases are further divided into their children for the successive steps. This process continues until the limit of the Recursive Segmentation Algorithm is reached, or the size of the segments becomes the size of pixels. These steps are carried out using the sub-Algorithm Disease Feature detection (Algorithm 4). 
\begin{figure*}[h]
	\centering
	\includegraphics[width=.9\linewidth]{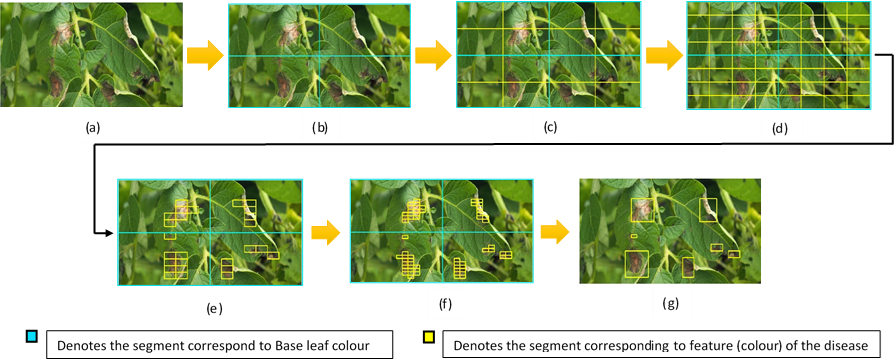}
	\caption{a)Input Image b)The image is divided based on the base leaf colour, and child segments are used for the next layer.	c)Each segment corresponding to the base leaf colour is checked for disease using Neural Network. d)The child segment of the segments corresponding to  the detected disease is checked for  the base colour. e) and f)Recursive steps- the child segments of the segments corresponding to the detected disease are checked for feature colour (Late blight colour). g)The child segments are merged using BFS. Each merged segment is checked for a particular disease.}
	\label{FIG:disease_detection_process}
\end{figure*}
\begin{algorithm}
    \caption{Disease Feature Detection}
    \label{alg:diseasefeaturedetection}
    \begin{algorithmic}[1]
        \Require{FeatureDict, Depth Count, Limit\_Dictionary}
        \For{feature in FeatureDict}
            \If {feature $\neq$ base\_colour}
                \State Initialize new coordinates as an empty list
                \If{Depth Count $>$ Limit\_Dictionary[feature]}
                    \For{segment in FeatureDict[feature]}
                        \State Break the segment into four child segments and check for a feature (colour, texture etc.) corresponding to the disease in each of the segments.
                        \State Append segments which have the feature to the new coordinates
                    \EndFor
                    \State Clear the FeatureDict[feature] list
                    \State FeatureDict[feature] $\leftarrow$ new coordinates
                \EndIf
            \EndIf
        \EndFor
        \State Calculate the width and height of one segment
        \Ensure{FeatureDict, width, height}
    \end{algorithmic}
\end{algorithm}
The value of \texttt{Base\_Colour\_Threshold} depends on the size of the input image. For the plant village dataset, the threshold is kept at 0. The threshold can be increased proportionally to the size of the input image. The threshold can still be kept for larger images at 0, leading to faster results but fewer regions detected. Thus, a trade-off can be achieved between speed and accuracy depending on the detection task. 
\subsection{Disease Feature Grouping}
Once the diseased segments are detected using the hierarchical feature detection, the segments with a side/vertex in common are merged to obtain the resulting region. The lists of segments corresponding to the disease are retrieved using a tree search algorithm (BFS in this case), and a bounding box is constructed. This bounding box defines the ROI (region of interest) for the affected areas in the plant.   Any random diseased segment may be selected as the starting point for this retrieval. The pseudocode of the same is given in the Feature Grouping Algorithm ~\ref{alg:featuregrouping}. Once the grouping is complete, the algorithm outputs the diseases and the corresponding regions of the diseases. Figure \ref{FIG:disease_detection_process} depicts the steps involved in the disease detection and localisation process.
\begin{algorithm}
    \caption{Feature Grouping}
    \label{alg:featuregrouping}
    \begin{algorithmic}[1]
        \Require{Start Node, List of Segments of a feature, ROI}
        \If{Start Node in List}
            \If{ROI is empty}
                \State ROI $\leftarrow$ Start Node
            \Else
                \State Check and update vertices of ROI to include all vertices of Start Node
            \EndIf
            \State Remove Start Node from List of segments
            \State Iterate over List of Segments, store segments overlapping or sharing a side or vertex with ROI in a neighbours list
            \If{neighbours list is not empty}
                \For{neighbour in neighbours list}
                    \State Call this algorithm recursively with a neighbour as new Start Node
                \EndFor
            \EndIf
        \EndIf
        \Ensure{ROI}
    \end{algorithmic}
\end{algorithm}
\vspace{-10pt} 
\subsection{Localization}
\begin{figure}[H]
	\begin{subfigure}[b]{0.45\linewidth}
		\includegraphics[width=.8\linewidth]{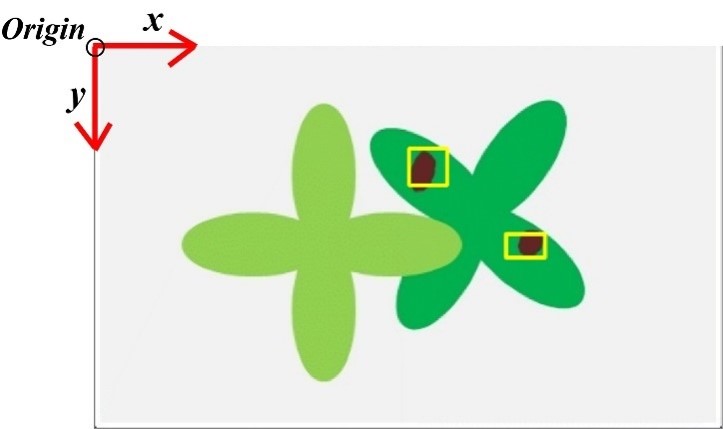}
		\caption{Schematic of the image coordinate system used by OpenCV in a sample test case.}
	\label{FIG:coordinate_system}
	\end{subfigure}
    \begin{subfigure}[b]{0.45\linewidth}
	    \centering
		\includegraphics[width=.8\linewidth]{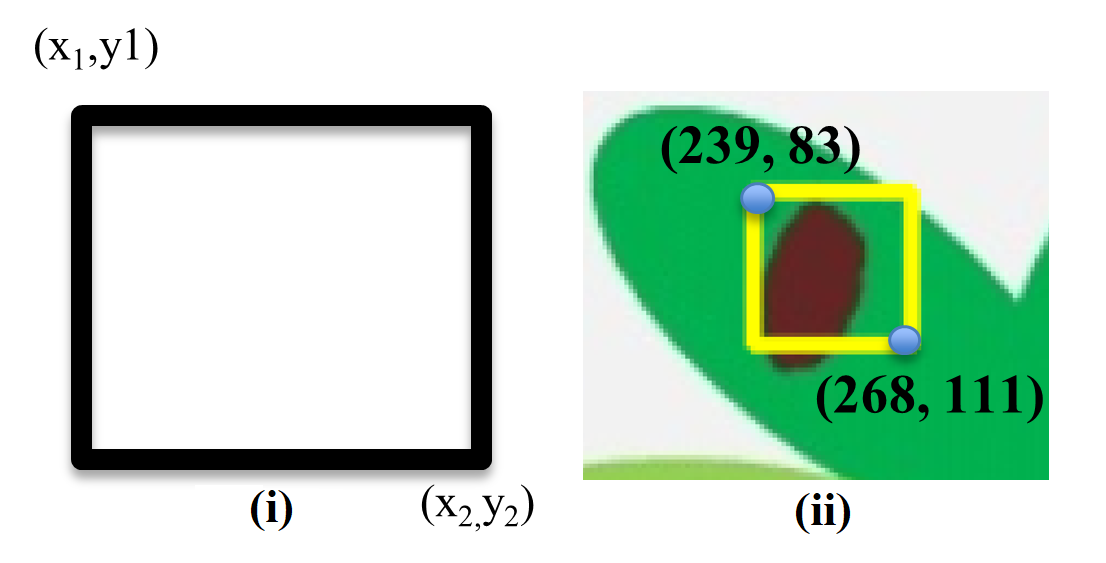}
		\caption{(i) An example of a bounding box and its corresponding coordinates. (ii)Close view of a deformed area and how the bounding box is quantified.}
	    \label{FIG:bounding_box}
    \end{subfigure}
    \caption{Disease Localization}
\end{figure}

After the segments of the diseases are grouped using Feature Grouping, the location of each region is obtained. The location is quantified using pixel coordinates. The coordinate system OpenCV uses is shown in Figure \ref{FIG:coordinate_system}. The origin is taken as the left top corner of the image, with the x-axis towards the right side and the y-axis downward. Each side of the pixel denotes one unit. The bounding boxes enclosing a disease-affected area are defined by opposite diagonal vertices as shown in Figure \ref{FIG:bounding_box}(i). The entire algorithm gives the output as a dictionary. The dictionary's key is the diseases, and their corresponding values are an array of bounding boxes, where each box is denoted by $[y_1, x_1, x_2, y_2]$. Figure \ref{FIG:bounding_box} shows an example of the localization of diseases found in Figure \ref{FIG:coordinate_system}. The image used in Figure \ref{FIG:coordinate_system} has a size of $(478 \times 296)$, and thus the output of the algorithm for the patch shown in Figure \ref{FIG:bounding_box}(ii) is $[83, 239, 111, 268]$

\section{Results and Discussion}
\label{section:result}
\begin{figure}[!ht]
    \begin{subfigure}[b]{0.24\linewidth}
	    \includegraphics[width=1\linewidth]{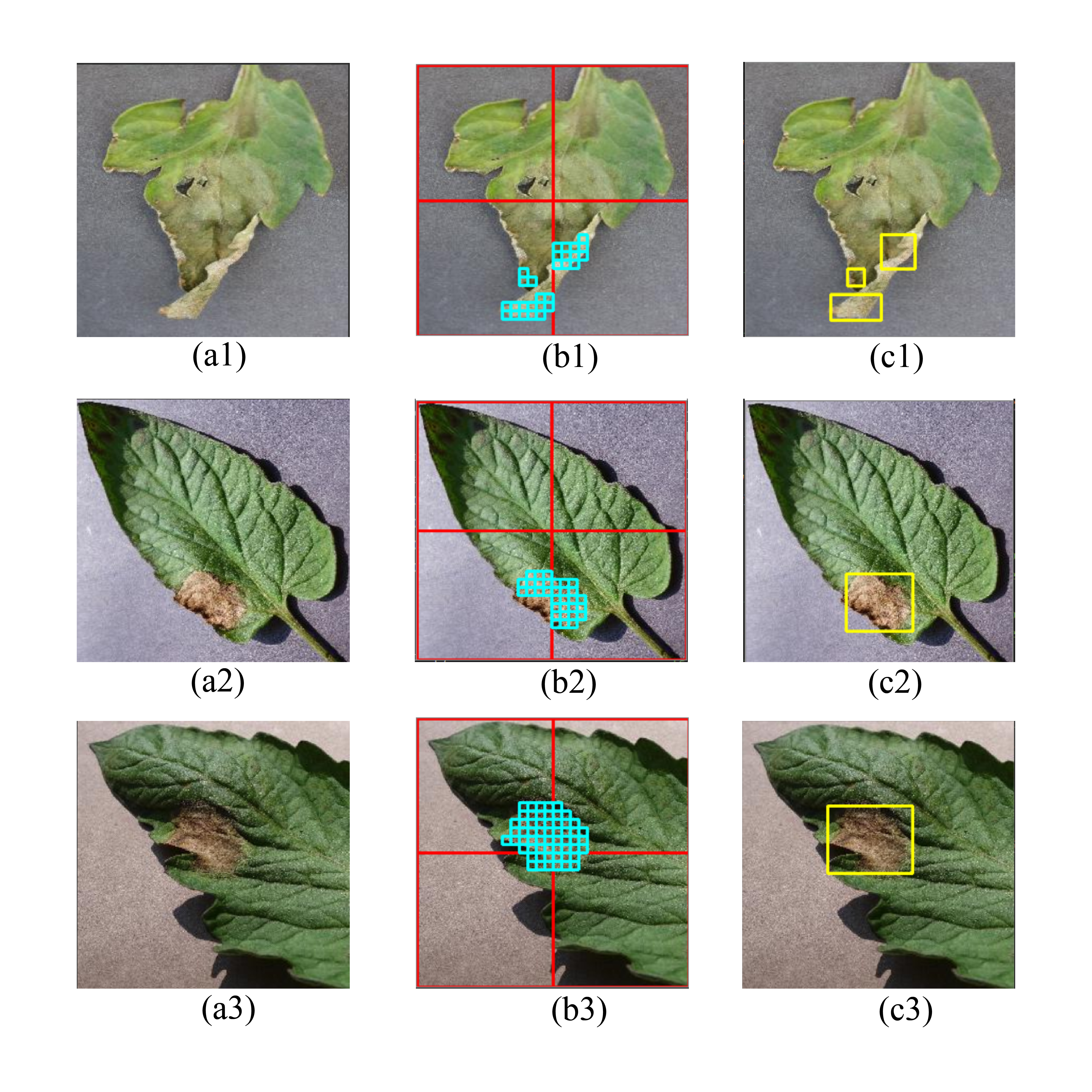}
        \caption{\label{FIG:Late_blight_tomato}Late blight in tomato leaves}
    \end{subfigure}
    \begin{subfigure}[b]{0.24\linewidth}
	    \includegraphics[width=1\linewidth]{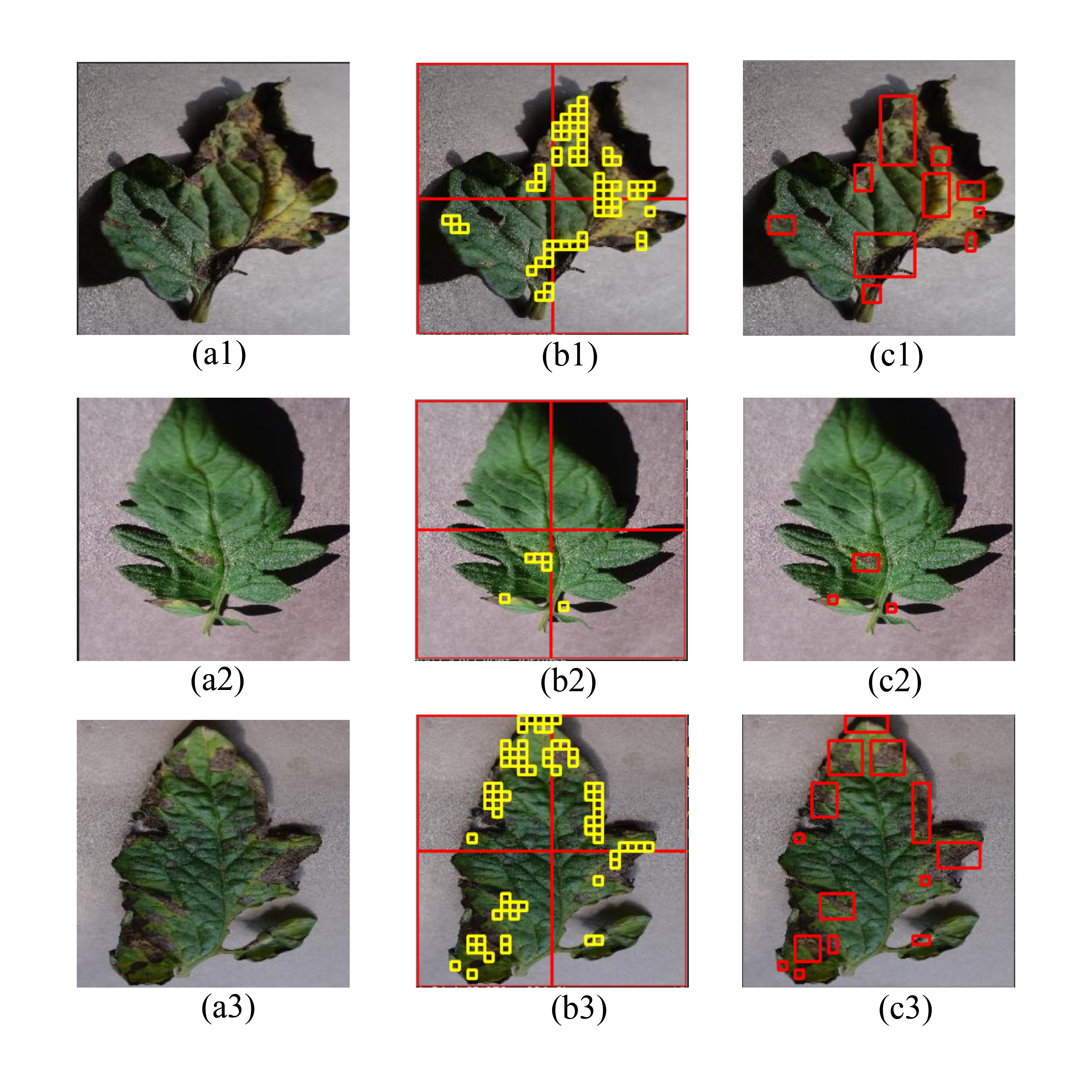}
	    \caption{\label{FIG:Early_blight_tomato}Early blight in tomato leaves}
    \end{subfigure}
    \begin{subfigure}[b]{0.24\linewidth}	
	    \includegraphics[width=1\linewidth]{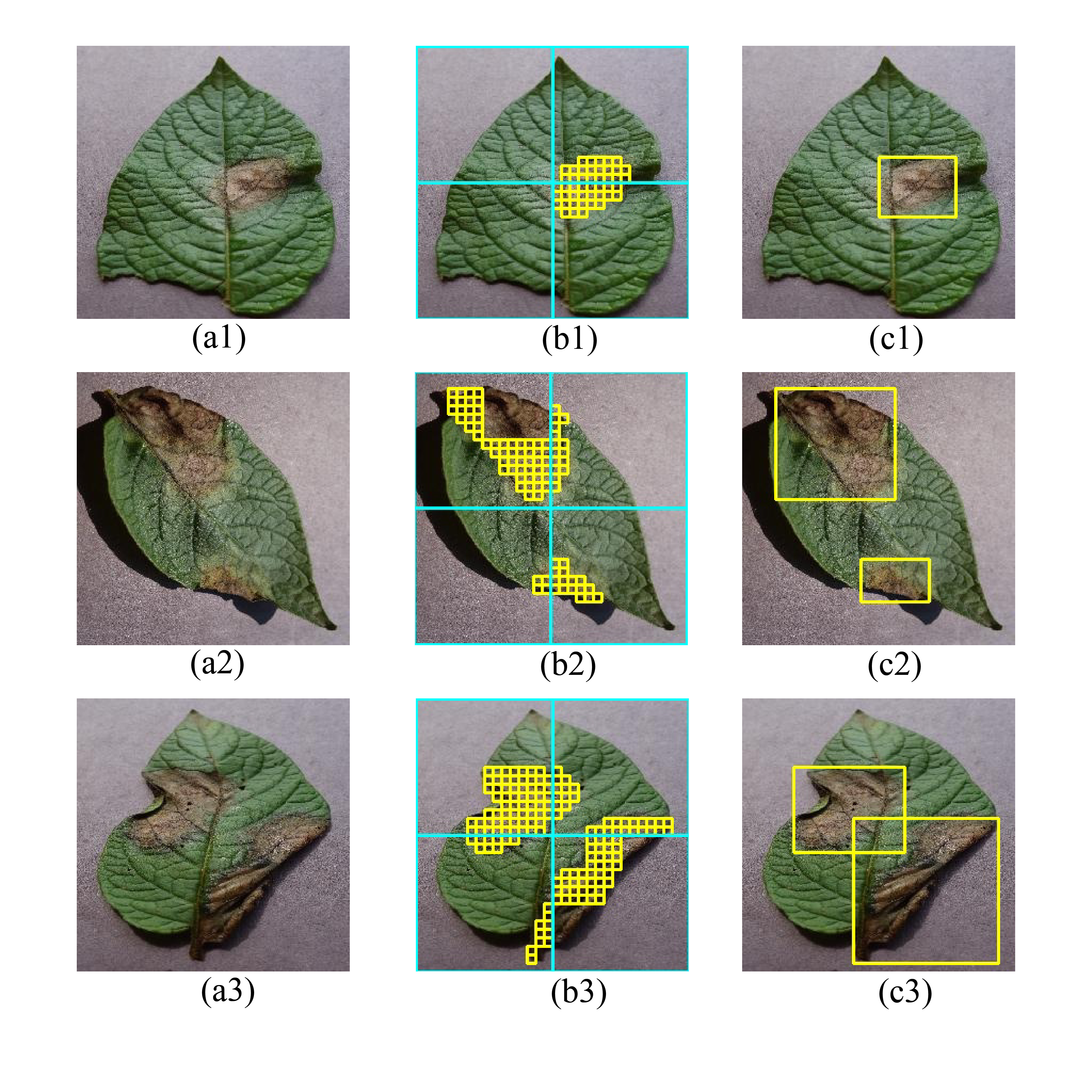}
	    \caption{\label{FIG:Late_blight_potato}Late blight in potato leaves}
    \end{subfigure}
    \begin{subfigure}[b]{0.24\linewidth}
	    \includegraphics[width=1\linewidth]{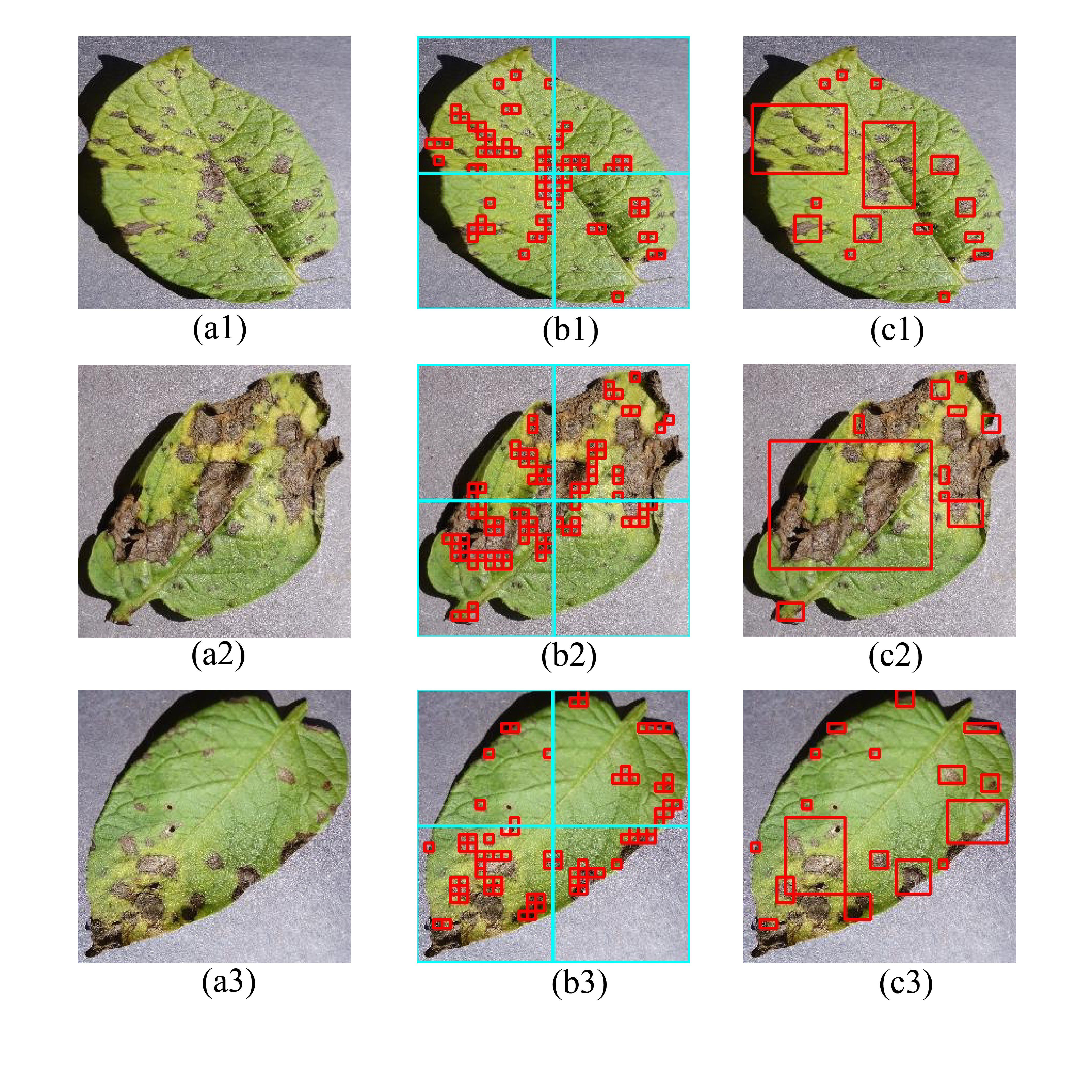}
	    \caption{\label{FIG:Early_blight_potato}Early blight in potato leaves}
    \end{subfigure}
    \caption{Disease Detection in Potato and Tomato Leaves}
\end{figure}

\begin{figure}[h]
    \begin{subfigure}{.5\linewidth}
    \centering
	\includegraphics[width=1\linewidth]{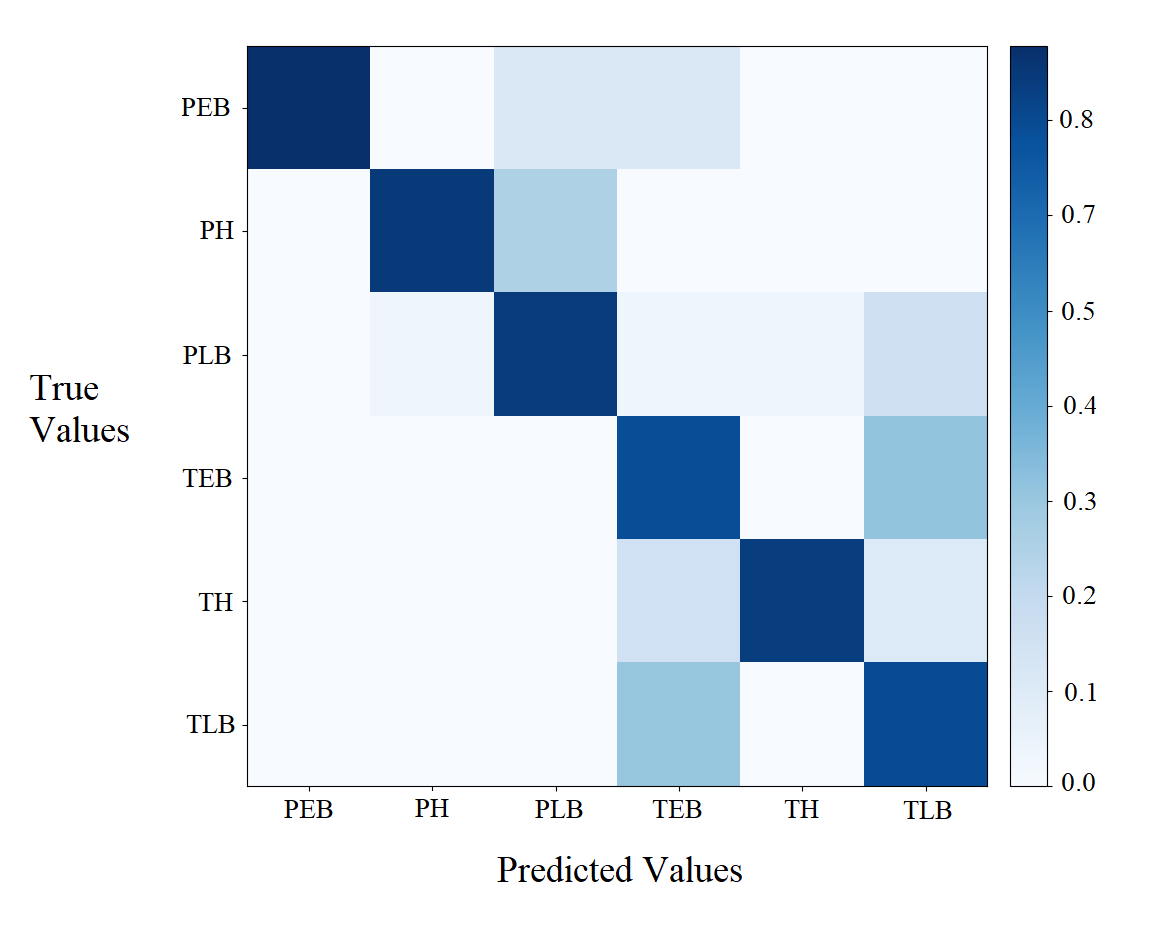}
	\caption{}
	\label{FIG:Confusion_matrix}
	\end{subfigure}
    \begin{subfigure}{.5\linewidth}
	\centering
	\includegraphics[width=1\linewidth]{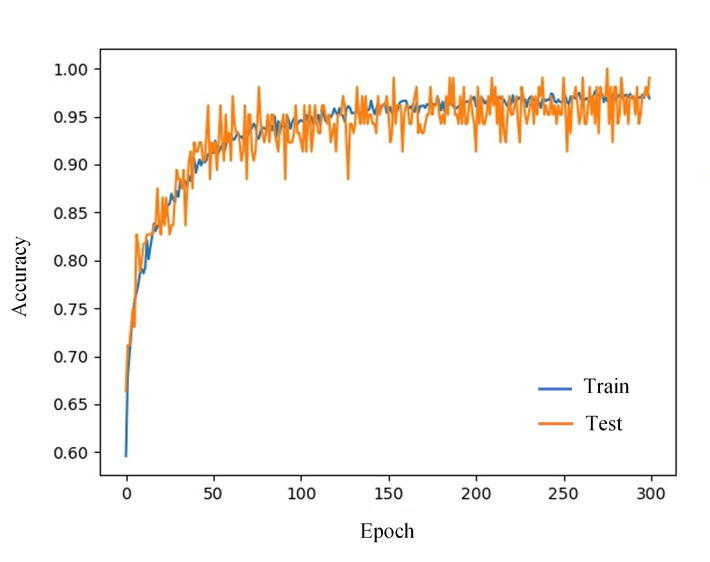}
	\caption{}
	\label{FIG:training}
	\end{subfigure}
    \caption{(a) Confusion Matrix for Potato Early Blight(PEB), Potato healthy(PH), Potato Late Blight(PLB), Tomato Late Blight(TEB),Tomato Healthy(TH), Tomato Late Blight(TLB)  (b)Training accuracy curve  for proposed model}
\end{figure}

The study utilized a combined dataset comprising 10,000 images, with the majority sourced from dataset collected at Chandrasekhar Azad Agriculture University in Kanpur, India, supplemented by samples from the publicly available Plant Village dataset. The proposed algorithm is tested on four disease classes, i.e., early blight and late blight, for two plants, potato and tomato.  The images in the data set are captured at different light exposure conditions. The proposed algorithm performs training on three classes, normal leaf, late blight, and early blight for each plant. The accuracy curve of the proposed Xception model is shown in Figure \ref{FIG:training}. Due to the sheer complexities of patterns in each class, especially in terms of infection status and scope, the system tends to be confused across many classes resulting in lower accuracy. Figure \ref{FIG:Early_blight_potato} and \ref{FIG:Late_blight_potato} show early and late blight in potato leaves. Figure \ref{FIG:Early_blight_tomato} and \ref{FIG:Late_blight_tomato} show the final detection results for early blight and late blight in tomato leaves. Figure \ref{FIG:Confusion_matrix} shows the associated confusion matrix for training across all disease classes. Depending on the outcomes, the classifier's performance can be visually evaluated, which helps to decide which classes and characteristics are typically recognized by the network's neurons. Confusion between classes may be reduced by fine-tuning the conditioning parameters corresponding to a specific disease, thereby increasing the accuracy.
The precision, recall and F1 score for four disease classes viz. Potato Late Blight (PLB), Potato Early Blight (PEB), Tomato Late Blight (TLB), and Tomato Early Blight(TEB) are shown in Table 1. 

The versatility of the proposed approach stems from the robustness of the method for real-time detection on high-resolution images, thereby preventing the loss of details arising from pre-processing techniques. 
\begin{table}[h!]
\label{tab:F1_score}
\caption{Performance Metrics Comparison}
\centering
\begin{tabular}{|l|l|l|l|l|}
\hline
\rowcolor[HTML]{C0C0C0} 
\textbf{Disease} & \textbf{Precision} & \textbf{Recall/Sensitivity} & \textbf{F1 Score} & \textbf{Specificity} \\ \hline
PLB              & 0.9664             & 0.725                       & 0.8288            & 0.9823               \\ \hline
PEB              & 0.7869             & 0.8421                      & 0.8136            & 0.9747               \\ \hline
TLB              & 0.843              & 0.8718                      & 0.8571            & 0.9634               \\ \hline
TEB              & 0.642              & 0.9231                      & 0.7579            & 0.9615               \\ \hline
\end{tabular}%
\end{table}
Table \ref{tab:Computationaltime} compares computational time for four disease classes on different GPUs. From this, it may be concluded that the proposed algorithm can run on the average system class for real-time processing; hence, it may be extended to embedded systems for real-time applications.
\begin{table}[h!]
\caption{Computational Time in Different GPUs}
\label{tab:Computationaltime}
\centering
\begin{tabular}{|c|c|c|c|c|}
\hline
\rowcolor[HTML]{C0C0C0} 
\textbf{\textbf{\begin{tabular}[c]{@{}c@{}}GPU  \\ Configuration/Parameter\end{tabular}}} &
  \textbf{\textbf{PEB}} &
  \textbf{\textbf{PLB}} &
  \textbf{\textbf{TEB}} &
  \textbf{\textbf{TLB}} \\ \hline
\cellcolor[HTML]{FFFFFF}\textbf{\begin{tabular}[c]{@{}c@{}}NVIDIA GeForce \\ GTX 1660 Ti\end{tabular}} & 0.063s & 0.057s & 0.054s & 0.050s \\ \hline
\cellcolor[HTML]{FFFFFF}\textbf{\begin{tabular}[c]{@{}c@{}}GT GeForce \\ GT 1030\end{tabular}}         & 0.117s & 0.111s & 0.125s & 0.126s \\ \hline
\cellcolor[HTML]{FFFFFF}\textbf{\begin{tabular}[c]{@{}c@{}}GT GeForce \\ GT 1050Ti\end{tabular}}       & 0.078s & 0.087s & 0.069s & 0.065s \\ \hline
\end{tabular}%
\end{table}
A comprehensive analysis of different algorithms, with a specific emphasis on evaluating their precision, accuracy, and extendibility. The algorithms under consideration do not include information regarding their processing speed, thus making it hard to compare them in this aspect. The findings of this comparative investigation are succinctly presented in Table \ref{tab:comparative_study}.
The proposed algorithm stands out because it can analyse larger images without scaling, which frequently removes small but potentially important details. This feature is essential for diagnosing diseases because these specifics may be useful. Other algorithms, such as AgroAid, iDohan, Ipedia, PlantBuddy, MS-Dnet, CROPCARE, and ToLeD, on the other hand, frequently demand that images be resized to 512x512 pixels, potentially losing crucial small features.

The algorithm performs well, with precision and accuracy values of 0.8 and 0.96 respectively. AgroAid and similar programmes demonstrate high precision and accuracy but might have trouble preserving fine details when resizing larger images.

The algorithm also exhibits remarkable extensibility, enabling customization to detect diseases in different plant species. In contrast, some algorithms, like iDohan, are only applicable to terrestrial plants.

As a result, the algorithm is a useful tool for diagnosing diseases because it combines precision, accuracy, and the capacity to process larger images without losing detail. The particular requirements of the task at hand should be considered when choosing an algorithm.
\begin{table}[H]
\caption{Precision, Accuracy, Speed and Extendibility Study}
\label{tab:comparative_study}
\resizebox{\columnwidth}{!}{%
\begin{tabular}{|l|l|l|l|}
\hline
\rowcolor[HTML]{C0C0C0} 
Algorithm  & Precision & Accuracy & Extendibility                                        \\ \hline
Proposed Algorithm &
  0.8 &
  0.96 &
  \begin{tabular}[c]{@{}l@{}}Currently 4 disease classes and highly extensible, \\  can be easily adapted to detect diseases in  other plant species\end{tabular} \\ \hline
AgroAid    & 0.99      & 0.99     & Can detect 39 disease classes                        \\ \hline
iDohan     & 0.96      & NP       & Currently only for terrestrial plants                \\ \hline
Ipedia     & NP        & 0.93     & Can detect Paddy Disease and 4 other disease classes \\ \hline
PlantBuddy & NP        & 0.97     & Can detect 26 different diseases                     \\ \hline
MS-Dnet    & 0.87      & 0.98     & Can detect 16 disease classes of Paddy and Corn      \\ \hline
CROPCARE   & NP        & 0.96     & Can detect 14 species with 38 disease classes        \\ \hline
ToLeD      & NP        & 0.912    & Can detect Tomato Leaf Disease                       \\ \hline
\end{tabular}%
}
\end{table}
*where NP= Not Provided
\section{Conclusion}
\label{section:conclusion}
This work uses a robust hybrid algorithm for detecting plant diseases in real-time. The technique introduces a practical and applicable solution for the classification and localization of diseases in plants, distinguishing itself from other methods for plant disease classification. The advantages of traditional image processing algorithms (Conditioned Quad-Tree segmentation) and the Deep Neural Network based techniques (modified Xception model) are utilized to design an efficient disease detection and localisation method.

Early Blight and Late Blight detection corresponding to potato and tomato crops demonstrate the proposed technique's performance. The overall F1 score depicting the accuracy was computed to be 0.79, which is considerably higher for practical applications. Improvement in detection accuracy may be achieved by expanding the training data set corresponding to disease classes to incorporate the wide variability of the disease features. This defines the future scope of the proposed algorithm.   
The significantly low computational requirement and faster processing time allow it to be readily deployed in platforms like drones, robots for precision agriculture and mobile app-based services for real-time crop disease identification. The proposed algorithm may be extended to various crops and develop a framework for a general-purpose crop disease detector.     

The present study employed a dataset comprising of individual plant leaves and plants that were gathered within a controlled laboratory setting and limited environmental variations. Nevertheless, the outcomes of our investigation indicate that the present dataset might not comprehensively encompass the intricacies inherent in real-life situations, given that variables such as fluctuations in lighting and diverse conditions in the field can exert a substantial influence on the integrity of image data.

It is imperative to employ a dataset replicating real-world scenarios to enhance the precision of the algorithm's evaluation. This methodology would facilitate the implementation of thorough examination and authentication of the algorithm within circumstances akin to its proposed utilisation, thereby augmenting our discoveries' overall transferability and practicality.

The scarcity of comprehensive natural data poses a significant challenge in applying artificial intelligence (AI) to agricultural issues. The outcomes obtained from numerous openly accessible datasets containing plant images, such as the extensively utilised PlantVillage dataset, might not precisely depict the algorithm's efficacy under real-world circumstances, primarily because these datasets primarily concentrate on controlled laboratory environments that employ artificial illumination.

The potential enhancement of the study lies in the retraining of the integrated model through the utilisation of a semi-supervised learning methodology, thereby capitalising on the acquisition of more precise user inputs derived from real-world scenarios. The utilisation of this particular approach has the potential to effectively establish a connection between outcomes derived from laboratory settings and their practicality in real-world scenarios.
\section{Acknowledgment}
This research was supported by the Department of Science and Technology, Government of India, through project number DST/ME/2020009. 

\bibliographystyle{unsrt}  
\bibliography{references}

\appendix
\section{Appendix-I}
\subsection{Precision}
Precision is how often the model is accurate when it predicts positive. Low precision indicates that there is a high false positives rate.

\begin{equation*}
Precision =\frac{true\ positive}{true\ positive\ + false\ positive}
\end{equation*}

\subsection{Recall or Sensitivity}
The recall is what proportion of actual positives was classified correctly, Low recall indicates a high false negatives rate.

\begin{equation*}
Recall = \frac{true\ positive}{ true\ positive\ + false\ negative}
\end{equation*}
\subsection{F1 score}
F1 is an overall measure of a model’s accuracy that combines precision and recall. A high F1 score means one has low false positives and false negatives.

\begin{equation*}
F1 = 2 *\frac{precision * recall}{precision + recall}  
\end{equation*}

\subsection{Specificity}
Specificity is the method used to measure a model's ability to predict true negatives for each accessible category. 
\begin{equation*}
Specificity= \frac{true\ negative}{true\  negative+false\ positive}     
\end{equation*}

\end{document}